\documentclass[conference]{IEEEtran}
\IEEEoverridecommandlockouts
\usepackage{cite}
\usepackage{amsmath,amssymb,amsfonts}
\usepackage{algorithm}
\usepackage[noend]{algorithmic}
\usepackage{graphicx}
\usepackage{textcomp}
\usepackage{xcolor}
\usepackage[a4paper, total={184mm,239mm}]{geometry}
\def\BibTeX{{\rm B\kern-.05em{\sc i\kern-.025em b}\kern-.08em
    T\kern-.1667em\lower.7ex\hbox{E}\kern-.125emX}}
\usepackage{tikz}
\newcommand*\circled[1]{\tikz[baseline=(char.base)]{
		\node[shape=circle,draw,inner sep=0.2pt] (char) {#1};}}

\usetikzlibrary{tikzmark}
\tikzset{circledColor/.style={circle,draw,inner sep=0.1em,line width=0.04em}}

\usepackage{enumitem}
\usepackage{multirow}
\usepackage{fancyhdr}
\fancypagestyle{firstpage}{
  \fancyhf{}
  \chead{Accepted at the Design, Automation and Test in Europe Conference (DATE), April 20th - 22nd, 2026 in Verona, Italy.}
  \cfoot{\thepage}
}

\begin{document}

%%%%%%%%%%%%%%%%%%%%%%%%%%%%%%%%%%%%%%%%%%%%%%%%%%%%%%%%%%%%%%%%%%%%%%%%%%%%%%%%%%
%%%%%%%%%%%%%%%%%%%%%%%%%%%%%%%%%%%%%%%%%%%%%%%%%%%%%%%%%%%%%%%%%%%%%%%%%%%%%%%%%%
\title{\huge QSLM: A Performance- and Memory-aware Quantization Framework with Tiered Search Strategy for Spike-driven Language Models
\vspace{-0.4cm}
}

\author{\IEEEauthorblockN{Rachmad Vidya Wicaksana Putra, Pasindu Wickramasinghe, Muhammad Shafique}
\IEEEauthorblockA{\textit{eBRAIN Lab, New York University (NYU) Abu Dhabi, Abu Dhabi, United Arab Emirates} \\
\{rachmad.putra, pmw6287, muhammad.shafique\}@nyu.edu}
\vspace{-1cm}
}

\maketitle
\pagestyle{plain}
\thispagestyle{firstpage}

%%%%%%%%%%%%%%%%%%%%%%%%%%%%%%%%%%%%%%%%%%%%%%%%%%%%%%%%%%%%%%%%%%%%%%%%%%%%%%%%%%
%%%%%%%%%%%%%%%%%%%%%%%%%%%%%%%%%%%%%%%%%%%%%%%%%%%%%%%%%%%%%%%%%%%%%%%%%%%%%%%%%%
\begin{abstract}
Large Language Models (LLMs) have been emerging as prominent AI models for solving many natural language tasks due to their high performance (e.g., accuracy) and capabilities in generating high-quality responses to the given inputs.
However, their large computational cost, huge memory footprints, and high processing power/energy make it challenging for their embedded deployments. 
Amid several tinyLLMs, recent works have proposed spike-driven language models (SLMs) for significantly reducing the processing power/energy of LLMs. 
However, their memory footprints still remain too large for low-cost and resource-constrained embedded devices. 
Manual quantization approach may effectively compress SLM memory footprints, but it requires a huge design time and compute power to find the quantization setting for each network, hence making this approach not-scalable for handling different networks, performance requirements, and memory budgets.
To bridge this gap, we propose \textit{QSLM}, a novel framework that performs automated quantization for compressing pre-trained SLMs, while meeting the performance and memory constraints. 
To achieve this, QSLM first identifies the hierarchy of the given network architecture and the sensitivity of network layers under quantization, then employs a tiered quantization strategy (e.g., global-, block-, and module-level quantization) while leveraging a multi-objective performance-and-memory trade-off function to select the final quantization setting. 
Experimental results indicate that our QSLM reduces memory footprint by up to 86.5\%, reduces power consumption by up to 20\%, maintains high performance across different tasks (i.e., by up to 84.4\% accuracy of sentiment classification on the SST-2 dataset and perplexity score of 23.2 for text generation on the WikiText-2 dataset) close to the original non-quantized model while meeting the performance and memory constraints. 
Hence, QSLM framework advances the efforts in enabling efficient design automation for embedded implementation of SLMs. 
\end{abstract}

\begin{IEEEkeywords}
Large Language Models (LLMs), Spike-driven Language Models (SLMs), Quantization, Memory Footprint, Embedded Systems, Design Automation.
\end{IEEEkeywords}

%%%%%%%%%%%%%%%%%%%%%%%%%%%%%%%%%%%%%%%%%%%%%%%%%%%%%%%%%%%%%%%%%%%%%%%%%%%%%%%%%%
%%%%%%%%%%%%%%%%%%%%%%%%%%%%%%%%%%%%%%%%%%%%%%%%%%%%%%%%%%%%%%%%%%%%%%%%%%%%%%%%%%
\vspace{-0.2cm}
\section{Introduction}
\label{Sec_Intro}
\vspace{-0.1cm}

Transformer-based networks~\cite{Ref_Vaswani_Attention_NIPS17} have achieved state-of-the-art performance (e.g., accuracy) in diverse machine learning (ML)-based applications, including solving diverse natural language tasks~\cite{Ref_Zhao_LLMsurvey_arXiv23, Ref_Minaee_LLMsurvey_arXiv24, Ref_Chang_LLMsurvey_TIST24, Ref_Han_SurveyViT_TPAMI22, Ref_Khan_SurveyViT_CSUR22, Ref_Putra_QSViT_IJCNN25}.
In recent years, transformer-based large language models (LLMs) have demonstrated significant improvements in extending the capabilities of natural language models~\cite{Ref_Zhao_LLMsurvey_arXiv23, Ref_Minaee_LLMsurvey_arXiv24, Ref_Chang_LLMsurvey_TIST24}, thereby making it possible to produce high-quality language-based understanding and responses to the given inputs.
Therefore, their adoption in resource-constrained embedded devices is highly in demand and actively being pursued for enabling customized and personalized systems~\cite{Ref_ElMir_LLM4Healthcare_ICIPCW24}. 
However, their large compute costs, huge memory footprints, and high processing power/energy make it challenging for their embedded deployments.

\begin{figure}[t]
\centering
\includegraphics[width=\linewidth]{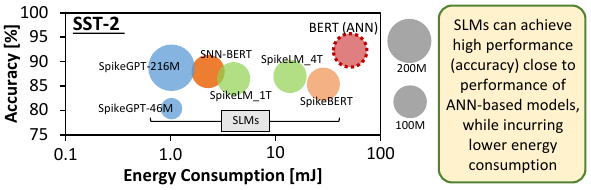}
\vspace{-0.7cm}
\caption{Current trends of performance (i.e., accuracy), number of weight parameters (note, M denotes millions [10$^6$] of parameters), and energy consumption of SLMs~\cite{Ref_Xing_SpikeLM_ICML24, Ref_Bal_SpikingBERT_AAAI24, Ref_Su_SNNBERT_NeuNet24, Ref_Xing_SpikeLLM_ICLR24, Ref_Chu_SpikeGPT_TMLR24} on the sentiment analysis task with the SST-2 dataset~\cite{Ref_Wang_GLUE_ICLR2019}.}
\label{Fig_Trends}
\vspace{-0.5cm}
\end{figure}

In the other domain, the advancements of spiking neural networks (SNNs) have demonstrated promising power/energy-efficient alternative to artificial neural network (ANN) algorithms, because of their sparse spike-driven operations~\cite{Ref_Bartolozzi_EmbodiedNeuroIntel_Nature22}\cite{Ref_Putra_SNNonCNP_IJCNN25}.
Therefore, recent works leveraged spike-driven operations for LLMs to reduce the processing power/energy requirements, i.e., so-called \textit{Spike-driven Language Models (SLMs)}; see Fig.~\ref{Fig_Trends}. 
However, their memory footprints are still too large for embedded deployments.
To reduce memory footprints of spike-driven models, quantization is one of the prominent methods~\cite{Ref_Putra_FSpiNN_TCAD20, Ref_Rathi_PruneQuantizeSNN_TCAD18, Ref_Putra_QSpiNN_IJCNN21}, because it can effectively reduce memory footprints with slightly yet acceptable accuracy degradation.
However, manually determining an appropriate quantization setting for any given SLM requires huge design time and large power/energy consumption. 
Therefore, this approach is laborious and \textit{not scalable} for compressing different SLMs for different possible performance and memory constraints.
Moreover, existing ANN quantization frameworks cannot be employed directly for SLMs due to the fundamental differences in synaptic and neuronal operations between ANNs and SNNs.

Such conditions lead us to \textbf{the research problem} targeted in this paper, i.e., \textit{how can we efficiently quantize any given pre-trained SLM, while maintaining high performance (e.g., accuracy) and meeting the memory constraint?} 
A solution to this problem may advance the design automation for efficient embedded implementation of SLMs.

%%%%%%%%%%%%%%%%%%%%%%%%%%%%%%%%%%%%%%%%%%%
\subsection{State-of-the-art of SLMs and Their Limitations}
\label{Sec_Intro_SOTA}

SLM development is a relatively new research avenue, hence the state-of-the-art works still focus on achieving high performance (e.g., accuracy), such as SpikeBERT~\cite{Ref_Lv_SpikeBERT_arXiv24}, SpikingBERT~\cite{Ref_Bal_SpikingBERT_AAAI24}, SNN-BERT~\cite{Ref_Su_SNNBERT_NeuNet24}, SpikeLM~\cite{Ref_Xing_SpikeLM_ICML24}, SpikeLLM~\cite{Ref_Xing_SpikeLLM_ICLR24}, and SpikeGPT~\cite{Ref_Chu_SpikeGPT_TMLR24}.
Specifically, spike-driven BERTs~\cite{Ref_Bal_SpikingBERT_AAAI24}\cite{Ref_Su_SNNBERT_NeuNet24}\cite{Ref_Lv_SpikeBERT_arXiv24} leverage BERT networks from ANN domain and apply spiking neuronal dynamics on them, while employing different techniques, such as knowledge distillation~\cite{Ref_Bal_SpikingBERT_AAAI24} and input coding enhancements~\cite{Ref_Su_SNNBERT_NeuNet24}. 
SpikeLM~\cite{Ref_Xing_SpikeLM_ICML24} and SpikeLLM~\cite{Ref_Xing_SpikeLLM_ICLR24} target to scale up spiking neuronal dynamics to large models (e.g., up to 70 billions of weight parameters for SpikeLLM). 
Meanwhile, SpikeGPT~\cite{Ref_Chu_SpikeGPT_TMLR24} targets at reducing the computational complexity in SLMs by replacing the spike-driven transformer modules with the spike-driven receptance weighted key value (SRWKV) modules, while maintaining the high performance. 
\textit{These state-of-the-art highlight that the efforts for quantizing SLMs have not been comprehensively explored}.

%%%%%%%%%%%%%%%%%%%%%%%%%%%%%%%%%%%%%%%%%%%
\subsection{A Case Study and Associated Research Challenges}
\label{Sec_Intro_Challenges}

\begin{figure}[t]
\centering
\includegraphics[width=\linewidth]{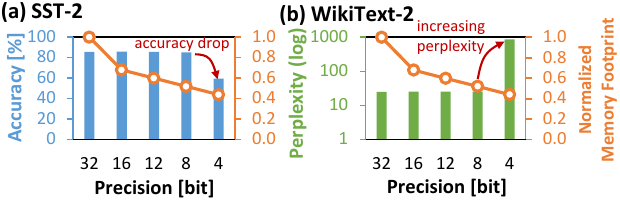}
\vspace{-0.7cm}
\caption{Performance profiles of the pre-trained SpikeGPT-216M after uniformly quantizing its weight parameters in its attention blocks across different precision levels for different tasks: (a) sentiment classification on the SST-2 dataset~\cite{Ref_Wang_GLUE_ICLR2019}, and (b) perplexity on the WikiText-2 dataset~\cite{Ref_Merity_WikiText_ICLR17}. 
Note, a lower perplexity score represents a better text generation performance.}
\label{Fig_Observe}
\vspace{-0.5cm}
\end{figure}

To show the potentials and challenges in quantizing SLMs, we perform an experimental case study.
Here, we apply uniform quantization on the weight parameters of the pre-trained SpikeGPT-216M~\cite{Ref_Chu_SpikeGPT_TMLR24} with the same precision level across its attention blocks, then employ the quantized model for solving the sentiment analysis on the SST-2 dataset~\cite{Ref_Wang_GLUE_ICLR2019} and evaluating the perplexity on the WikiText-2 dataset~\cite{Ref_Merity_WikiText_ICLR17}. 
Further details of experiments are provided in Sec.~\ref{Sec_EvalMethod}, and the experimental results are presented in Fig.~\ref{Fig_Observe}.
These results show that, the post-training quantization (PTQ) scheme leads to significant memory reduction, while preserving high performance (e.g., accuracy and perplexity) when employed with appropriate quantization.
Otherwise, it leads to notable performance degradation. 

Furthermore, these observations also expose several research challenges, as follows.  
\begin{itemize}[leftmargin=*]
    \item Quantization process should handle different network complexity levels (e.g., number of layers) efficiently.
    \item Quantization process should be able to meet different possible performance (e.g., accuracy) and memory constraints, thus making it practical for diverse applications. 
    \item Quantization process should minimize manual intervention to increase its scalability for handling different networks, performance requirements, and memory budgets.
\end{itemize}

%%%%%%%%%%%%%%%%%%%%%%%%%%%%%%%%%%%%%%%%%%%
\subsection{Our Novel Contributions}
\label{Sec_Intro_Novelty}

\begin{figure}[h]
\vspace{-0.3cm}
\centering
\includegraphics[width=\linewidth]{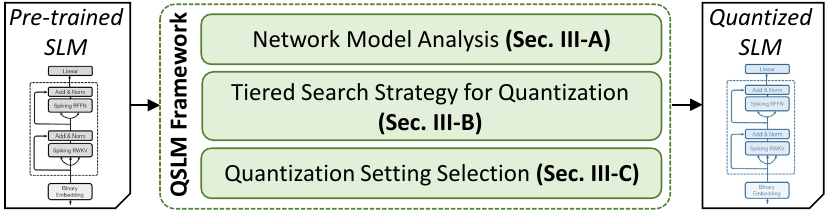}
\vspace{-0.7cm}
\caption{Overview of our novel contributions.}
\label{Fig_Novelty}
\vspace{-0.4cm}
\end{figure}

To address the targeted problem and research challenges, we propose \textit{\textbf{QSLM}, a novel framework that performs automated \underline{Q}uantization for compressing pre-trained \underline{S}pike-driven \underline{L}anguage \underline{M}odel (SLM) to meet the performance (e.g., accuracy) and memory constraints}.
To achieve this, QSLM performs the following key steps; see an overview in Fig.~\ref{Fig_Novelty}. 
\begin{itemize}[leftmargin=*]
    \item \textbf{Network Model Analysis (Sec.~\ref{Sec_QSLM_NetAnalysis}):} 
    It aims to identify the structure of the given pre-trained model, determine the network hierarchy to be considered for quantization search, and investigate the sensitivity of each block of the network under quantization on the performance (e.g., accuracy). 
    \item \textbf{Tiered Search Strategy for Quantization (Sec.~\ref{Sec_QSLM_Search}):} 
    It aims to perform automated quantization and evaluation for the model candidates under different phases (e.g., global-, block-, and module-level quantization, subsequently) based on the network hierarchy and the sensitivity analysis, while considering the performance and memory constraints.
    \item \textbf{Quantization Setting Selection (Sec.~\ref{Sec_QSLM_QSelect}):} 
    It selects the final quantization setting from the candidates by leveraging our trade-off function that quantifies the candidates' benefits based on their performance and memory footprint.  
\end{itemize}

\textbf{Key Results:}
We implement the QSLM framework using PyTorch and then run it on the Nvidia RTX A6000 multi-GPU machine. 
Experimental results show that QSLM provides effective quantization settings for SLMs. 
It saves by up to 86.5\% of memory footprint, reduces by up to 20\% of power consumption, and maintain high performance across different tasks (i.e., by up to 84.4\% accuracy of sentiment classification on the SST-2 and 23.2 perplexity score of text generation on the WikiText-2) close to the non-quantized model, while meeting the performance and memory constraints. 
These results show the potential of QSLM framework for enabling embedded implementation of SLMs. 

%%%%%%%%%%%%%%%%%%%%%%%%%%%%%%%%%%%%%%%%%%%%%%%%%%%%%%%%%%%%%%%%%%%%%%%%%%%%%%%%%%
%%%%%%%%%%%%%%%%%%%%%%%%%%%%%%%%%%%%%%%%%%%%%%%%%%%%%%%%%%%%%%%%%%%%%%%%%%%%%%%%%%
\section{Background}
\label{Sec_Back}

\textbf{SNNs:}
An SNN model design typically encompasses spiking neurons, network architecture, neural/spike coding, and learning rule~\cite{Ref_Putra_FSpiNN_TCAD20}\cite{Ref_Mozafari_SpykeTorch_FNINS19}.
Recent SNN developments in software~\cite{roy2019towards, Ref_Rathi_SNNsurvey_CSUR23, Ref_Putra_TopSpark_IROS23, Ref_Chowdhury_TemporalPruneSNN_ECCV22, Ref_Putra_SpikeNAS_TAI25} and hardware~\cite{Ref_Akopyan_TrueNorth_TCAD15, Ref_Roy_PEASE_ISLPED17, Ref_Davies_Loihi_MM18, Ref_Neckar_Braindrop_IEEE19, Ref_Frenkel_ODIN_TBCAS19, Ref_Frenkel_MorphIC_TBCAS19, Ref_SynSense_DYNAP, Ref_BrainChip_Akida} have advanced the practicality of SNNs for diverse ultra-low power/energy application use-cases. 

\textbf{SLMs:}
Recently, several state-of-the-art SLMs have been proposed in the literature, such as SpikeBERT~\cite{Ref_Lv_SpikeBERT_arXiv24}, SpikingBERT~\cite{Ref_Bal_SpikingBERT_AAAI24}, SNN-BERT~\cite{Ref_Su_SNNBERT_NeuNet24}, SpikeLM~\cite{Ref_Xing_SpikeLM_ICML24}, SpikeLLM~\cite{Ref_Xing_SpikeLLM_ICLR24}, and SpikeGPT~\cite{Ref_Chu_SpikeGPT_TMLR24}. 
In this work, we consider SpikeGPT as the potential model candidate for embedded systems since it offers competitive performance with the lowest energy consumption due to its reduced computational complexity; see Fig.~\ref{Fig_Trends}. 
Specifically, SpikeGPT replaces traditional self-attention mechanism with Spiking Receptance Weighted Key Value (SRWKV) and Spiking Receptance Feed-Forward Networks (SRFFN). 

\begin{figure}[h]
\vspace{-0.2cm}
\centering
\includegraphics[width=\linewidth]{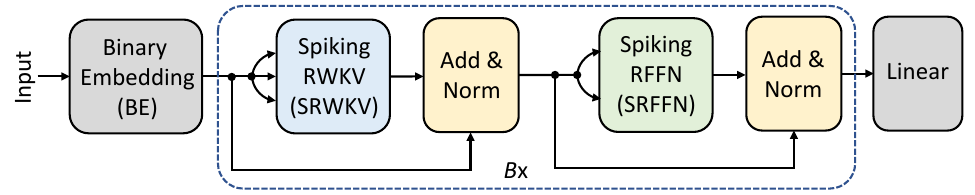}
\vspace{-0.7cm}
\caption{Overview of the SpikeGPT architecture. $B$ is the number of attention blocks. For instance, the pre-trained SpikeGPT-216M has $B$=18 blocks~\cite{Ref_Chu_SpikeGPT_TMLR24}.}
\label{Fig_SpikeGPT}
\vspace{-0.2cm}
\end{figure}

%%%
\begin{figure*}[t]
\centering
\includegraphics[width=\linewidth]{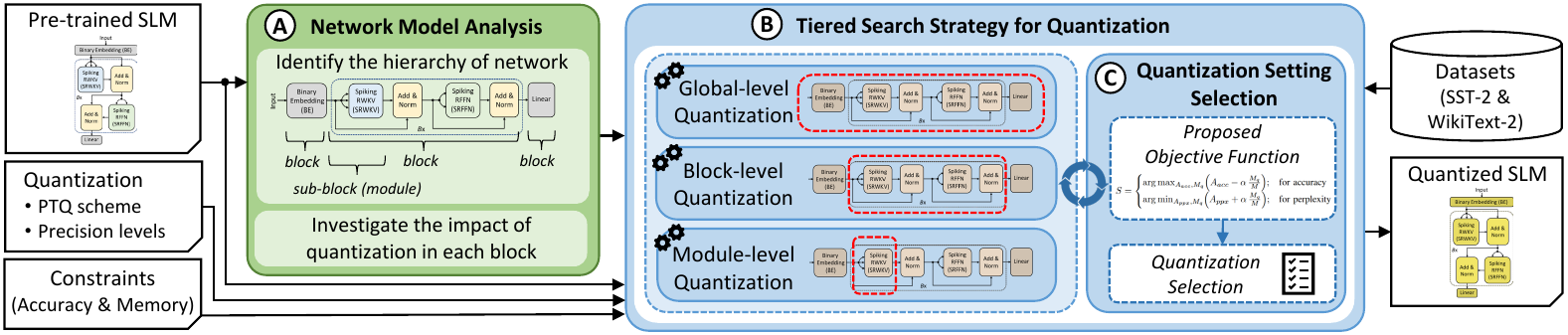}
\vspace{-0.7cm}
\caption{Our QSLM framework showing its key steps: network model analysis, tiered search strategy for quantization, and quantization setting selection.}
\label{Fig_QSLM}
\vspace{-0.4cm}
\end{figure*}
%%%

SRWKV leverages element-wise products rather than matrix-matrix multiplication, hence reducing the computational cost of the attention mechanism.
Meanwhile, SRFFN is employed to replace the conventional feed-forward network (FFN) with a spiking-compatible version.
Its network architecture is illustrated in Fig.~\ref{Fig_SpikeGPT} and summarized in Table~\ref{Tab_SpikeGPT}.
If the data have been processed through all network layers, the model either employs a classification head for natural language understanding (NLU) or a generation head for natural language generation (NLG).

\begin{table}[t]
\caption{The architectural hierarchy of the SpikeGPT-216M~\cite{Ref_Chu_SpikeGPT_TMLR24}. 
Note, the attention parameters in SRWKV include $K$, $V$, and $R$, which denote Key, Value, and Receptance, respectively.}
\vspace{-0.1cm}
\scriptsize
\centering
\begin{tabular}{|c|c|c|c|c|c|}
\hline
  \textbf{Block} & \begin{tabular}[c]{@{}c@{}} \textbf{Sub-Block} \\ \textbf{(Module)} \end{tabular} & \begin{tabular}[c]{@{}c@{}} \textbf{Number of} \\ \textbf{Parameters} \end{tabular} & \textbf{Quantity} & \begin{tabular}[c]{@{}c@{}} \textbf{Total Number of} \\ \textbf{Parameters} \end{tabular}\\ \hline \hline
  Input & \begin{tabular}[c]{@{}c@{}} Embedding \\ Layer Norm. \end{tabular} & \begin{tabular}[c]{@{}c@{}} 38.6M \\ \; 1.5K \end{tabular} & 1 & \begin{tabular}[c]{@{}c@{}} 38.6M \end{tabular} \\  \hline
  Attention & \begin{tabular}[c]{@{}c@{}} Layer Norm. \\ SRWKV  \\ SRFFN \end{tabular} & \begin{tabular}[c]{@{}c@{}} \;\;\;\; 3K \\ \;\; 2.4M \\ \;\; 5.3M \end{tabular} & 18 & \begin{tabular}[c]{@{}c@{}} 138.2M \end{tabular}\\  \hline
  Output & \begin{tabular}[c]{@{}c@{}} Layer Norm. \\ Head \end{tabular} & \begin{tabular}[c]{@{}c@{}} \; 1.5K \\ 38.6M \end{tabular} & 1 & \begin{tabular}[c]{@{}c@{}} 38.6M \end{tabular} \\  \hline
\end{tabular}
\label{Tab_SpikeGPT}
\vspace{-0.4cm}
\end{table}

%%%%%%%%%%%%%%%%%%%%%%%%%%%%%%%%%%%%%%%%%%%
\subsection{SNN Quantization}
\label{Sec_Back_Quant}

There are two possible schemes for quantizing SNN models, namely \textit{Quantization-aware Training (QAT)} and \textit{Post-Training Quantization (PTQ)}~\cite{Ref_Putra_QSpiNN_IJCNN21}\cite{Ref_Krishnamoorthi_Whitepaper_arXiv18}. 
QAT quantizes an SNN model during the training phase based on the given precision level.
Meanwhile, PTQ quantizes the pre-trained SNN model with the given precision level.  
In this work, we consider the PTQ scheme since it avoids the expensive training costs, such as the computational time, memory, and power/energy consumption~\cite{Ref_Liu_SpinQuant_ICLR25}. 
To realize this, we employ the simulated quantization approach to enable fast design space exploration and provide representative results in performance (e.g., accuracy) and power/energy consumption saving~\cite{Ref_vanBaalen_SimQuantRealPower_CVPR22}.

%%%%%%%%%%%%%%%%%%%%%%%%%%%%%%%%%%%%%%%%%%%%%%%%%%%%%%%%%%%%%%%%%%%%%%%%%%%%%%%%%%
%%%%%%%%%%%%%%%%%%%%%%%%%%%%%%%%%%%%%%%%%%%%%%%%%%%%%%%%%%%%%%%%%%%%%%%%%%%%%%%%%%
\section{The QSLM Framework}
\label{Sec_QSLM}

We propose the novel QSLM framework to solve the targeted problem and related challenges, whose overview is presented in Fig.~\ref{Fig_QSLM}.
It employs \circled{A} \textit{network model analysis} to identify the model structure and identify its block sensitivity under quantization, \circled{B} \textit{tiered search strategy} to systematically perform quantization on the model, and \circled{C} \textit{quantization setting selection} that considers performance and memory constraints.  
Details of its key steps are discussed in the following sub-sections.

%%%%%%%%%%%%%%%%%%%%%%%%%%%%%%%%%%%%%%%%%%%
\subsection{Network Model Analysis}
\label{Sec_QSLM_NetAnalysis}

\begin{figure}[t]
\centering
\includegraphics[width=\linewidth]{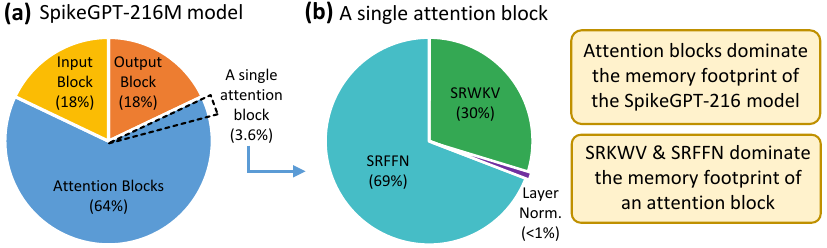}
\vspace{-0.7cm}
\caption{Proportion of the memory footprint for (a) the SpikeGPT-216M model with its blocks, and (b) a single attention block with its sub-blocks/modules.}
\label{Fig_PieSpikeGPT}
\vspace{-0.2cm}
\end{figure}

To perform effective quantization, it is important to apply appropriate precision levels on the weight parameters of the model.
Therefore, \textit{this step targets to understand the network structure of the model, identify its architectural hierarchy for quantization search, and investigate its block sensitivity under quantization on the performance}, through the following ideas.
\begin{itemize}[leftmargin=*]
    \item We identify blocks in the network model that can be quantized. 
    Typically, they are categorized as the \textit{input}, \textit{attention}, and \textit{output blocks}. 
    \item For each block, we identify the sub-blocks (modules) and the respective number of weights to estimate the memory saving potentials; see Table~\ref{Tab_SpikeGPT} and Fig.~\ref{Fig_PieSpikeGPT} for SpikeGPT-216M.
    \item Then, we investigate the block sensitivity under quantization by applying different precision levels to individual block and evaluating the performance (e.g., accuracy).
    It is useful for devising a suitable strategy for quantization search.
\end{itemize}

\begin{figure}[t]
\centering
\includegraphics[width=\linewidth]{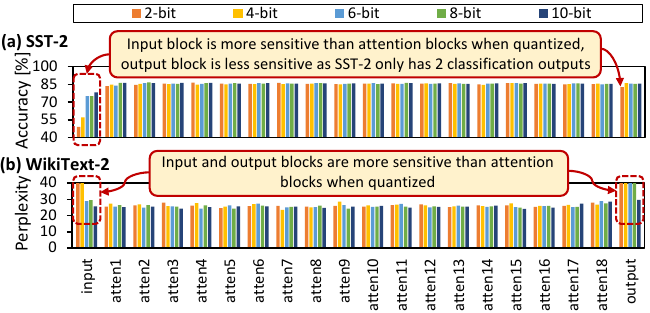}
\vspace{-0.8cm}
\caption{Results of block-wise quantization in SpikeGPT-216M across different precision levels for (a) accuracy of sentiment classification task on the SST-2 dataset, and (b) perplexity of text generation task on the WikiText-2 dataset.
Note, a lower perplexity score means a better performance of text generation.}
\label{Fig_BlockSensitivity}
\vspace{-0.4cm}
\end{figure}

For instance, we conduct experiments that apply different weight precision levels on each block of the SpikeGPT-216M for sentiment analysis on the SST-2 and text generation on the WikiText-2. 
Experimental results are presented in Fig.~\ref{Fig_BlockSensitivity}, from which we make the following key observations. 

\begin{itemize}[leftmargin=*]
    \item The input and output blocks are more sensitive than the attention blocks, since the loss of information from quantization in these blocks lead to notable performance degradation.
    Therefore, \textit{the input and output blocks should be carefully quantized to maximize memory reduction while ensuring high performance (e.g., accuracy)}.
    \item The attention blocks are less sensitive than the input/output block. 
    Considering that the attention blocks dominate the memory footprint, quantizing them potentially lead to significant memory reduction. 
    Therefore, \textit{quantizing the attention blocks is beneficial to achieve significant memory reduction}.
\end{itemize}
These observations are then leveraged in Sec.~\ref{Sec_QSLM_Search} to enable automated quantization process.

%%%%
\begin{algorithm}[t]
\caption{Tiered search strategy for quantization}
\label{Alg_QuantSearch}
\footnotesize
\begin{algorithmic}[1]
\renewcommand{\algorithmicrequire}{\textbf{INPUT:}}
\renewcommand{\algorithmicensure}{\textbf{OUTPUT:}}
\REQUIRE \textbf{(1)} Pre-trained model ($Net$), its performance ($P$) and memory footprint ($M$); 
\textbf{(2)} Pre-defined bit precision levels $b$: $b\in$ $\{16, 14, 12, ... 4\}$, and its number of precision levels ($N_b$); 
\textbf{(3)} Number of blocks in the model ($N_k$);
\textbf{(4)} Number of modules in the attention block ($N_m$);
\textbf{(5)} Constraints: performance constraint ($const_A$), and memory constraint ($const_M$);
\ENSURE Quantized model ($Net_q$); \\
\smallskip
\textbf{BEGIN} \\
\smallskip
\textbf{Initialization}: \\
\STATE $c$ = 0;
\STATE $candQ[c,:,:]$ = 32; \\
\STATE $P$, $M$ = test($Net$, $candQ[c,:,:]$); \\
\STATE $cStat[c].perf$ = $P$; \\
\STATE $cStat[c].mem$ = $M$; \\
\smallskip
\textbf{Process}: \\
\smallskip
\underline{// Global-level quantization}
\FOR{($i$ = 0; $i<$ $N_b$; $i$++)}
  \STATE $c$ = $c$+1;
  \STATE $candQ[c,:,:]$ = $b[i]$; 
  \STATE $Net_{t}$ = quantize($Net$, $candQ[c,:,:]$);
  \STATE $cStat[c]$, $X$ = eval($Net_{t}$, $P$, $M$, $const_A$, $const_M$); \textcolor{teal}{// Alg.~\ref{Alg_Eval}}
  \IF{($X$ == `constraints are met')}
    \STATE $cStat[c].met$ = `true';
    \STATE $I_{last}$ = $i$;
  \ELSE 
    \STATE $cStat[c].met$ = `false';
    \STATE $I_{tmp}$ = $I_{last}$;
  \ENDIF
\ENDFOR

\smallskip
\underline{// Block-level quantization}
\FOR{($k$ = 0; $k<$ $N_k$; $k$++)}
  \FOR{($i$ = $I_{tmp}$; $i<$ $N_b$; $i$++)}
    \STATE $c$ = $c$+1;
    \STATE $candQ[c,k,:]$ = $b[i]$; 
    \STATE $Net_{t}$ = quantize($Net$, $candQ[c,:,:]$);
  \STATE $cStat[c]$, $X$ = eval($Net_{t}$, $P$, $M$, $const_A$, $const_M$); \textcolor{teal}{// Alg.~\ref{Alg_Eval}}
    \IF{($X$ == `constraints are met')}
      \STATE $cStat[c].met$ = `true';
      \STATE $I_{last2}[k]$ = $i$;
    \ELSE 
      \STATE $cStat[c].met$ = `false';
      \STATE $I_{tmp2}[k]$ = $I_{last2}[k]$;
    \ENDIF
  \ENDFOR
\ENDFOR

\smallskip
\underline{// Module-level quantization}
\FOR{($k$ = 1; $k<$ ($N_k$-1); $k$++)}
  \FOR{($m$ = 0; $m<$ $N_m$; $m$++)}
  \FOR{($i$ = $I_{tmp2}[k]$; $i<$ $N_b$; $i$++)}
    \STATE $c$ = $c$+1;
    \STATE $candQ[c,k,m]$ = $b[i]$; 
    \STATE $Net_{t}$ = quantize($Net$, $candQ[c,:,:]$);
  \STATE $cStat[c]$, $X$ = eval($Net_{t}$, $P$, $M$, $const_A$, $const_M$); \textcolor{teal}{// Alg.~\ref{Alg_Eval}}
    \STATE $cStat[c].score$ = $S_{tmp}$;
    \IF{($X$ == `constraints are met')}
      \STATE $cStat[c].met$ = `true';
      \STATE $I_{last3}[k,m]$ = $i$;
    \ELSE 
      \STATE $cStat[c].met$ = `false';
    \ENDIF
  \ENDFOR
  \ENDFOR
\ENDFOR
\smallskip
\STATE $cand_{fin}$ = select($candQ$, max($cStat[:].score$), $cStat[:].met$); \\
\STATE $Net_q$ = quantize($Net$, $cand_{fin}$);
\STATE \textbf{return} $Net_q$; \\
\textbf{END}
\end{algorithmic}
\end{algorithm}
\setlength{\textfloatsep}{4pt}
%%%%
%%%%
\begin{algorithm}[t]
\caption{Evaluation of the quantized model candidate}
\label{Alg_Eval}
\footnotesize
\begin{algorithmic}[1]
\renewcommand{\algorithmicrequire}{\textbf{INPUT:}}
\renewcommand{\algorithmicensure}{\textbf{OUTPUT:}}
\REQUIRE \textbf{(1)} Performance ($P$) and memory footprint ($M$) of the original non-quantized model;
\textbf{(2)} Input model ($Net_{tmp}$); 
\textbf{(3)} Constraints: performance (i.e., accuracy/perplexity) constraint ($const_A$), and memory constraint ($const_M$);
\textbf{(4)} Candidate index ($c$);
\ENSURE \textbf{(1)} Characteristics of the model candidates ($cStat$); \textbf{(2)} Status if constraints are met ($X$: `true'/`false'); \\
\smallskip
\textbf{BEGIN} \\
\smallskip
\textbf{Process}: \\
  \STATE $P_{tmp}$, $M_{tmp}$ = test($Net_{tmp}$);
  \STATE $S_{tmp}$ = calc\_score($P_{tmp}$, $M_{tmp}$); \textcolor{teal}{// Eq.~\ref{Eq_Score}}
  \STATE $X$ = check($P$, $M$, $const_P$, $const_M$, $P_{tmp}$, $M_{tmp}$); 
  \STATE $cStat[c].perf$ = $P_{tmp}$;
  \STATE $cStat[c].mem$ = $M_{tmp}$;
  \STATE $cStat[c].score$ = $S_{tmp}$;
\STATE \textbf{return} $cStat$, $X$; \\
\textbf{END}
\end{algorithmic}
\end{algorithm}
\setlength{\textfloatsep}{4pt}
%%%%

%%%%%%%%%%%%%%%%%%%%%%%%%%%%%%%%%%%%%%%%%%%
\subsection{Tiered Search Strategy for Quantization}
\label{Sec_QSLM_Search}

This step aims to enable an automated quantization process to maximize the memory reduction, while meeting both performance constraint ($const_A$) and memory constraint ($const_M$). 
To obtain this, \textit{we propose a tiered search strategy that applies a certain bit precision level ($b$) to the targeted weights from the highest-level network hierarchy to the lowest one (e.g., global-level, block-level, and module-level quantization, subsequently)}. 
Its key steps are described below (pseudocode in Alg.~\ref{Alg_QuantSearch} and~\ref{Alg_Eval}). 
\begin{itemize}[leftmargin=*]
    \item \textbf{Global-level quantization:} 
    We uniformly quantize all blocks in the model based on the pre-defined list of precision levels ($b$), such as $b\in$ $\{16, 14, 12, ..., 4\}$. 
    Here, we orderly apply $b$ value from the largest to the smallest ones, while evaluating if the quantized model meets both $const_{A}$ and $const_{M}$.   
    \begin{itemize}
        \item If both constraints are met, then the investigated precision level $b$ is recorded as the quantization candidate ($candQ$).
        \item If both constraints are not met, then the selected precision is set back to the last acceptable precision (from index-$I_{last}$ of list $b$). 
        Then, we move to \textit{block-level quantization}.
    \end{itemize}
    \item \textbf{Block-level quantization:} 
    We quantize each block in the model with lower precision than the previously applied one in the global-level step. 
    Then, we subsequently apply lower precision based on the list $b$, while performing evaluation. 
    \begin{itemize}
        \item If both constraints are met, then the investigated precision level $b$ is recorded as the setting for the respective block, and used to update the candidate $candQ$.
        \item If both constraints are not met, then the selected precision for the respective block is set back to the last acceptable precision level (from index-$I_{last2}$ of list $b$). 
        Afterward, we move to \textit{module-level quantization}.
    \end{itemize}
    \item \textbf{Module-level quantization:} 
    We quantize each module in the attention blocks with lower precision than the previously applied one in the block-level step. 
    We further apply lower precision based on the list $b$, while performing evaluation.
    \begin{itemize}
        \item If both constraints are met, then the investigated precision level $b$ is recorded as the quantization setting for the respective module, and used to update the candidate $candQ$.
        \item If both constraints are not met, then the precision level $b$ for the respective module is set back to the last acceptable precision level (from index-$I_{last3}$ of list $b$). 
    \end{itemize}
\end{itemize}

%%%%%%%%%%%%%%%%%%%%%%%%%%%%%%%%%%%%%%%%%%%
\subsection{Quantization Setting Selection}
\label{Sec_QSLM_QSelect}

The tiered search strategy may obtain multiple quantization candidates that meet $const_A$ and $const_M$.
To select the most appropriate solution, we quantify the benefit of the candidates considering their performance (e.g., accuracy) and memory saving, and then select the one with the highest score ($S$).
To do this, \textit{we propose a performance-and-memory trade-off function, that can be expressed as Eq.~\ref{Eq_Score}}. 
Here, $A_{acc}$ denotes accuracy for classification task and $A_{ppx}$ denotes perplexity for generation task; $M$ and $M_q$ denote memory footprints for the original non-quantized model and quantized model, respectively; and $\alpha$ denotes the user-defined adjustment factor.  
In the classification task, it aims to maximize the score $S$, that is proportional to the accuracy, since higher accuracy is better.
In the generation task, it aims to minimize the score $S$, since lower perplexity is better.
A candidate with larger memory than other candidates will penalize more the score $S$. 
Furthermore, perplexity score $A_{ppx}$ can be calculated using Eq.~\ref{Eq_Perplexity}, with $N_T$ is the number of words (tokens) in the sequence, and $P(w_i \mid w_{<i})$ is the model's predicted probability of word $w_i$ given the previous words.
\begin{equation}
    S  = 
    \begin{cases}
     \arg\max_{A_{acc}, M_q} \Bigl( A_{acc} - \alpha \, \frac{M_q}{M} \Bigr); & \text{for accuracy} \\
      \arg\min_{A_{ppx}, M_q} \Bigl( A_{ppx} + \alpha \, \frac{M_q}{M} \Bigr); & \text{for perplexity}
    \end{cases}
    \label{Eq_Score}
\end{equation}
\begin{equation}
    A_{ppx} = \exp\left(-\frac{1}{N_T} \sum_{i=1}^{N_T} \log P(w_i \mid w_{<i})\right) 
    \label{Eq_Perplexity}
\end{equation}

%%%%%%%%%%%%%%%%%%%%%%%%%%%%%%%%%%%%%%%%%%%%%%%%%%%%%%%%%%%%%%%%%%%%%%%%%%%%%%%%%%
%%%%%%%%%%%%%%%%%%%%%%%%%%%%%%%%%%%%%%%%%%%%%%%%%%%%%%%%%%%%%%%%%%%%%%%%%%%%%%%%%%
\section{Evaluation Methodology}
\label{Sec_EvalMethod}

\begin{figure}[t]
\centering
\includegraphics[width=\linewidth]{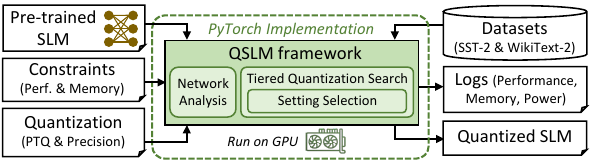}
\vspace{-0.7cm}
\caption{Experimental setup for the evaluation}
\label{Fig_ExpSetup}
\end{figure}

To evaluate the QSLM framework, we develop its PyTorch-based implementation, then run it on the Nvidia RTX A6000 multi-GPU machine; see Fig.~\ref{Fig_ExpSetup}.
For the baseline non-quantized model we consider the state-of-the-art pre-trained SpikeGPT-216M~\cite{Ref_Chu_SpikeGPT_TMLR24} that has been trained with 5B tokens from the OpenWebText dataset~\cite{pile}.
We use its publicly available pre-trained model and codes from the original authors, and then reproduce the fine-tuning and testing phases with their default hyperparameter settings on targeted tasks.
In the evaluation, we consider the following tasks: (1) a sentiment classification task on the SST-2 dataset~\cite{Ref_Wang_GLUE_ICLR2019}, and (2) a text generation task on the WikiText-2 dataset~\cite{Ref_Merity_WikiText_ICLR17}.
Under the baseline settings, we achieve accuracy of 85.7\% for the sentiment classification task, and perplexity score of 26.5 for the text generation task.
Here, we consider different sets of constraints to investigate the performance of QSLM under different constraint cases.
\begin{itemize}[leftmargin=*]
    \item In sentiment classification task, case-a1: $const_A$ = 2\% and $const_M$ = 400MB; case-a2: $const_A$ = 5\% and $const_M$ = 400MB; and case-a3: $const_A$ = 5\% and $const_M$ = 420MB.
    \item In text generation task, case-b1: $const_A$ = 1 and $const_M$ = 400MB; case-b2: $const_A$ = 4 and $const_M$ = 400MB; and case-b3: $const_A$ = 4 and $const_M$ = 420MB.
\end{itemize}
Note, $const_A$ denotes the maximum acceptable accuracy degradation or perplexity increase, while $const_M$ denotes the maximum acceptable memory footprint.
Furthermore, we also perform ablation study for investigating the impact of different $\alpha$ values with $\alpha$ $\in$ $\{0, 0.2, 0.4, 0.6, 0.8, 1\}$.
The experiments evaluate several metrics, such as accuracy for sentiment classification task, perplexity score for text generation task, memory footprint, and power consumption (using \textit{nvidia-smi} utility). 

%%%%%%%%%%%%%%%%%%%%%%%%%%%%%%%%%%%%%%%%%%%%%%%%%%%%%%%%%%%%%%%%%%%%%%%%%%%%%%%%%%
%%%%%%%%%%%%%%%%%%%%%%%%%%%%%%%%%%%%%%%%%%%%%%%%%%%%%%%%%%%%%%%%%%%%%%%%%%%%%%%%%%
\section{Results and Discussion}
\label{Sec_Results}

\begin{figure*}[t]
\centering
\includegraphics[width=\linewidth]{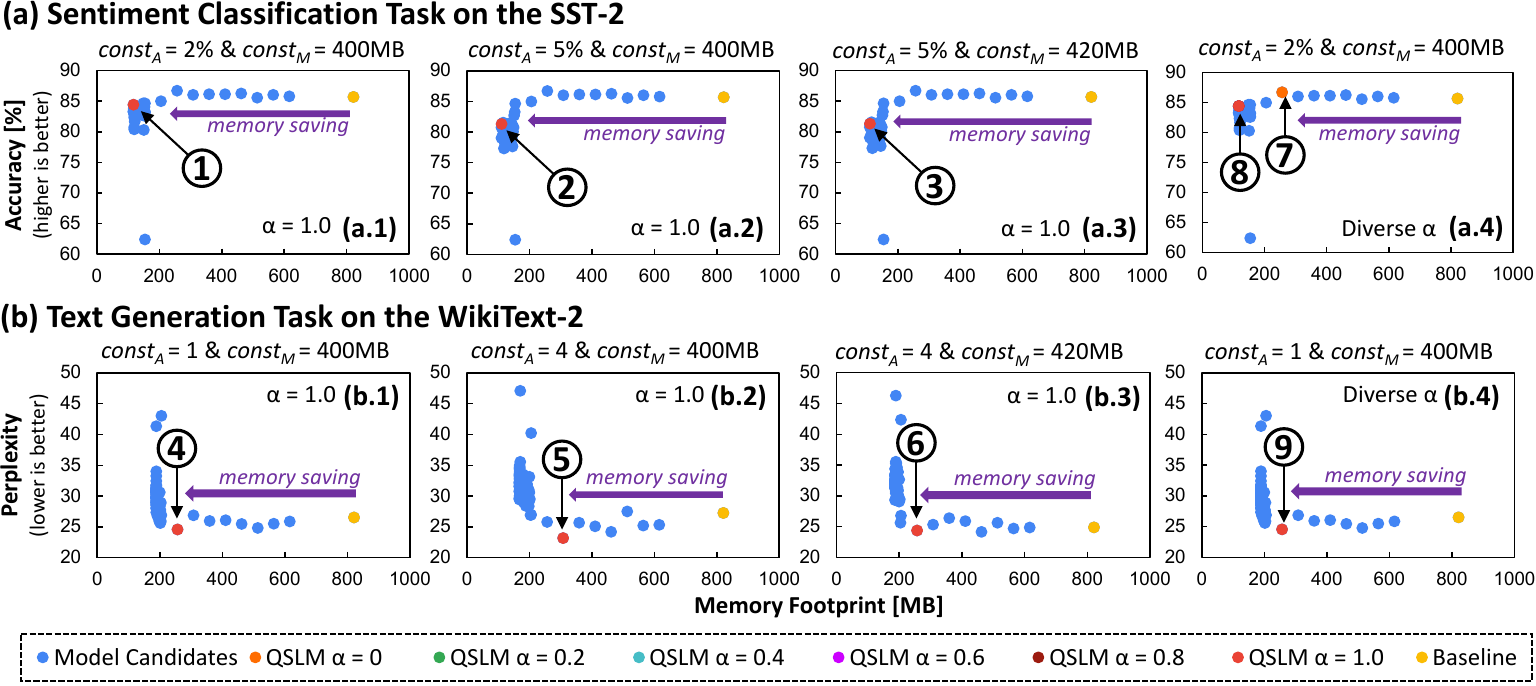}
\vspace{-0.7cm}
\caption{Experimental results of \textbf{(a)} sentiment classification task on the SST-2 for different sets of constraints (a1-a3) and diverse $\alpha$ (a4); and \textbf{(b)} text generation task on the WikiText-2 for different sets of constraints (b1-a3) and diverse $\alpha$ (b4).}
\label{Fig_Results}
\vspace{-0.3cm}
\end{figure*}

\begin{figure*}[t]
\centering
\includegraphics[width=0.95\linewidth]{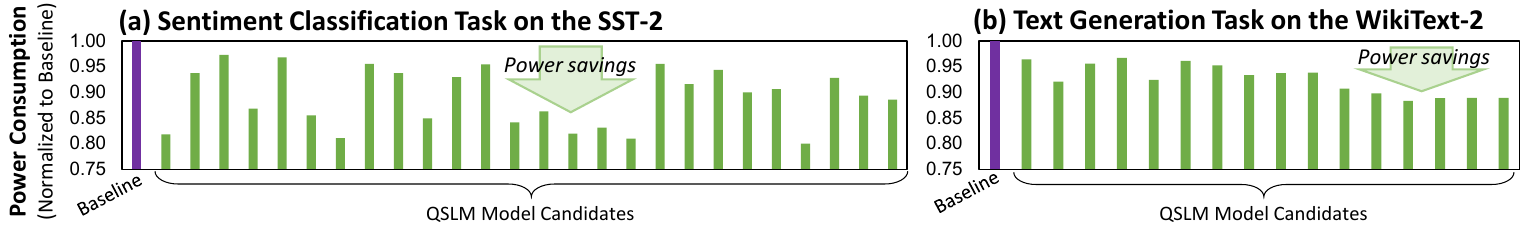}
\vspace{-0.3cm}
\caption{Experimental results of power consumption incurred by the baseline model and our QSLM model candidates that meet both constraints for \textbf{(a)} sentiment classification task on the SST-2, and \textbf{(b)} text generation task on the WikiText-2.}
\label{Fig_Results_Power}
\vspace{-0.4cm}
\end{figure*}

%%%%%%%%%%%%%%%%%%%%%%%%%%%%%%%%%%%%%%%%%%%
\subsection{Reducing Memory while Maintaining High Performance}
\label{Sec_Results_MemoryPerf}

Experimental results for sentiment classification task are provided in Fig.~\ref{Fig_Results}(a). 
These results show that, QSLM effectively reduces memory footprint of the baseline model across different scenarios (i.e., different sets of constraints and different $\alpha$ values), while meeting both accuracy and memory constraints. 
Our key observations are the following.
\begin{itemize}[leftmargin=*]
    \item For case-a1 in Fig.~\ref{Fig_Results}(a.1): QSLM achieves 84.4\% accuracy and reduces 85.7\% memory footprint; see \circled{1}. 
    \item For case-a2 in Fig.~\ref{Fig_Results}(a.2): QSLM achieves 81.3\% accuracy and reduces 86.5\% memory footprint; see \circled{2}. 
    \item For case-a3 in Fig.~\ref{Fig_Results}(a.3): QSLM achieves 81.3\% accuracy and reduces 86.5\% memory footprint; see \circled{3}.
\end{itemize}
Meanwhile, experimental results for text generation task are provided in Fig.~\ref{Fig_Results}(b). 
These results also show that, QSLM effectively reduces memory footprint of the baseline model across different scenarios (i.e., different sets of constraints and different $\alpha$ values), while meeting both perplexity and memory constraints. 
Our key observations are the following.
\begin{itemize}[leftmargin=*]
    \item For case-b1 in Fig.~\ref{Fig_Results}(b.1): QSLM achieves 24.6 perplexity score and reduces 68.7\% memory footprint; see \circled{4}.
    \item For case-b2 in Fig.~\ref{Fig_Results}(b.2): QSLM achieves 23.2 perplexity score and reduces 62.4\% memory footprint; see \circled{5}.
    \item For case-b3 in Fig.~\ref{Fig_Results}(b.3): QSLM achieves 24.4 perplexity score and reduces 68.7\% memory footprint; see \circled{6}.
\end{itemize}
These significant memory savings while preserving high accuracy or low perplexity can be obtained due to the systematic quantization approach in our QSLM framework.
Specifically, QSLM leverages the block sensitivity information from model analysis to guide the quantization search, then performs tiered search strategy to carefully apply different precision levels on different network blocks/modules, while ensuring the selected model candidates always meet the given constraints (i.e., $const_A$ and $const_M$) by leveraging a performance-and-memory trade-off function and the given constraints.   
%%%%%%%%%%%%%%%%%%%%%%%%%%%%%%%%%%%%%%%%%%%
\subsection{Reduction of Power Consumption}
\label{Sec_Results_Power}

Experimental results for power consumption of the baseline model and the QSLM model candidates that meet both performance and memory constraints are provided in Fig.~\ref{Fig_Results_Power}.
For the sentiment classification task, QSLM model candidates can reduce the power consumption by 2.6\%-20\%.
Meanwhile, for the text generation task, QSLM model candidates can reduce the power consumption by 3.2\%-11.6\%.
These power savings come from the reduction of precision levels in the weight parameters of the quantized models, thereby incurring lower computational and memory power to complete the processing, as compared to the baseline non-quantized model.
Furthermore, these results also demonstrate that, QSLM effectively optimizes power consumption, while meeting both performance and memory constraints (i.e., $const_A$ and $const_M$). 

%%%%%%%%%%%%%%%%%%%%%%%%%%%%%%%%%%%%%%%%%%%
\subsection{Impact of Different $\alpha$ Values on the Model Selection}
\label{Sec_Results_Alpha}

Experimental results for investigating the impact of different $\alpha$ values on the model selection are provided in Fig.~\ref{Fig_Results}(a.4) for sentiment classification task and Fig.~\ref{Fig_Results}(b.4) for text generation task. 
These results show that, different $\alpha$ values may lead to different model selection, as summarized below.
\begin{itemize}[leftmargin=*]
    \item In the sentiment classification task, $\alpha$ = 0 guides the QSLM search strategy to put the memory aspect as non-priority, and hence leading the selection process toward a model with higher accuracy and higher memory footprint, as pointed by \circled{7} in Fig.~\ref{Fig_Results}(a.4).  
    Meanwhile, the other investigated $\alpha$ values guide the QSLM search strategy to adjust the priority level of memory aspect proportional to the respective $\alpha$ value. 
    In this case study, QSLM search strategy selects a quantized model candidate that is pointed by \circled{8} in Fig.~\ref{Fig_Results}(a.4). 
    These results demonstrate that, our performance-and-memory trade-off function in QSLM effectively helps selection of quantized model based on the priority of memory footprint relative to performance (e.g., accuracy). 
    \item In the text generation task, all investigated $\alpha$ values lead the QSLM search strategy to select a model with 24 perplexity score and 68.7\% memory saving, as shown by \circled{9} in Fig.~\ref{Fig_Results}(b.4).  
    These results demonstrate that, there are some conditions that QSLM search strategy finds a relatively dominant quantized model in performance (e.g., perplexity), hence adjusting $\alpha$ with small values does not change the final selection for the quantized model. 
\end{itemize}
%

%%%%%%%%%%%%%%%%%%%%%%%%%%%%%%%%%%%%%%%%%%%%%%%%%%%%%%%%%%%%%%%%%%%%%%%%%%%%%%%%%%
%%%%%%%%%%%%%%%%%%%%%%%%%%%%%%%%%%%%%%%%%%%%%%%%%%%%%%%%%%%%%%%%%%%%%%%%%%%%%%%%%%
\section{Conclusion}
\label{Sec_Conclude}

In this paper, we propose the novel QSLM framework for performing automated quantization on the pre-trained SLMs. 
Our QSLM significantly reduces memory footprint by up to 86.5\%, decreases power consumption by up to 20\%, preserves high performance across different tasks (i.e., by up to 84.4\% accuracy for the SST-2 dataset and 23.2 perplexity score for the WikiText-2 dataset), while meeting the given accuracy and memory constraints.
These results also demonstrate that our QSLM successfully advances the efforts in enabling efficient design automation for embedded implementation of SLMs.

%%%%%%%%%%%%%%%%%%%%%%%%%%%%%%%%%%%%%%%%%%%%%%%%%%%%%%%%%%%%%%%%%%%%%%%%%%%%%%%%%%
%%%%%%%%%%%%%%%%%%%%%%%%%%%%%%%%%%%%%%%%%%%%%%%%%%%%%%%%%%%%%%%%%%%%%%%%%%%%%%%%%%
\section*{Acknowledgment}
This work was partially supported by the NYUAD Center for CyberSecurity (CCS), funded by Tamkeen under the NYUAD Research Institute Award G1104. 

%%%%%%%%%%%%%%%%%%%%%%%%%%%%%%%%%%%%%%%%%%%%%%%%%%%%%%%%%%%%%%%%%%%%%%%%%%%%%%%%%%
%%%%%%%%%%%%%%%%%%%%%%%%%%%%%%%%%%%%%%%%%%%%%%%%%%%%%%%%%%%%%%%%%%%%%%%%%%%%%%%%%%
\bibliographystyle{IEEEtran}
\bibliography{bibliography}

\end{document}